\documentclass[10pt,twocolumn,letterpaper]{article}

\usepackage[pagenumbers]{iccv} % To force page numbers, e.g. for an arXiv version

\usepackage{array,multirow,graphicx}
\newcolumntype{P}[1]{>{\centering\arraybackslash}p{#1}}
\usepackage{arydshln}
\definecolor{iccvblue}{rgb}{0.21,0.49,0.74}
\usepackage[pagebackref,breaklinks,colorlinks,allcolors=iccvblue]{hyperref}

\def\paperID{XXXXX} % *** Enter the Paper ID here
\def\confName{ICCV}
\def\confYear{2025}

\title{DHECA-SuperGaze: Dual Head-Eye Cross-Attention and Super-Resolution for Unconstrained Gaze Estimation}

\author{Franko Šikić \,\,\,\, Donik Vršnak \,\,\,\, Sven Lončarić\\
University of Zagreb Faculty of Electrical Engineering and Computing\\
Unska 3, Zagreb, Croatia\\
{\tt\small franko.sikic@fer.unizg.hr}
}

\begin{document}
\maketitle
\begin{abstract}
Unconstrained gaze estimation is the process of determining where a subject is directing their visual attention in uncontrolled environments.
Gaze estimation systems are important for a myriad of tasks such as driver distraction monitoring, exam proctoring, accessibility features in modern software, etc.
However, these systems face challenges in real-world scenarios, partially due to the low resolution of in-the-wild images and partially due to insufficient modeling of head-eye interactions in current state-of-the-art (SOTA) methods.
This paper introduces DHECA-SuperGaze, a deep learning-based method that advances gaze prediction through super-resolution (SR) and a dual head-eye cross-attention (DHECA) module.
Our dual-branch convolutional backbone processes eye and multiscale SR head images, while the proposed DHECA module enables bidirectional feature refinement between the extracted visual features through cross-attention mechanisms.
Furthermore, we identified critical annotation errors in one of the most diverse and widely used gaze estimation datasets, Gaze360, and rectified the mislabeled data.
Performance evaluation on Gaze360 and GFIE datasets demonstrates superior within-dataset performance of the proposed method, reducing angular error (AE) by 0.48° (Gaze360) and 2.95° (GFIE) in static configurations, and 0.59° (Gaze360) and 3.00° (GFIE) in temporal settings compared to prior SOTA methods. 
Cross-dataset testing shows improvements in AE of more than 1.53° (Gaze360) and 3.99° (GFIE) in both static and temporal settings, validating the robust generalization properties of our approach.
\end{abstract}    
\section{Introduction}
\label{sec:intro}

Appearance-based gaze estimation is the computational process of determining where a person directs their visual attention based on an image or video recording.
It has emerged as an important technology for understanding human behavior and enhancing human-machine interaction.
The ability to detect eye movement and where a subject is looking provides insight into cognitive processes, intentions, or possible distractions, enabling novel applications across diverse domains.
These application domains include exam proctoring systems that can detect off-screen glances to ensure academic integrity \cite{exam_proctoring}, or driver monitoring solutions that can identify distraction or fatigue \cite{driver_monitoring}. 
Furthermore, its applications extend into augmented human-computer interfaces, where real-time gaze tracking allows individuals with motor disabilities to more easily control their devices \cite{human_computer_gaze} and into consumer analytics, where predicting gaze-object interactions can reveal the subject's preferences in retail or advertising contexts \cite{retail_gaze}.

\begin{figure}
\centering
\includegraphics[width=\columnwidth]{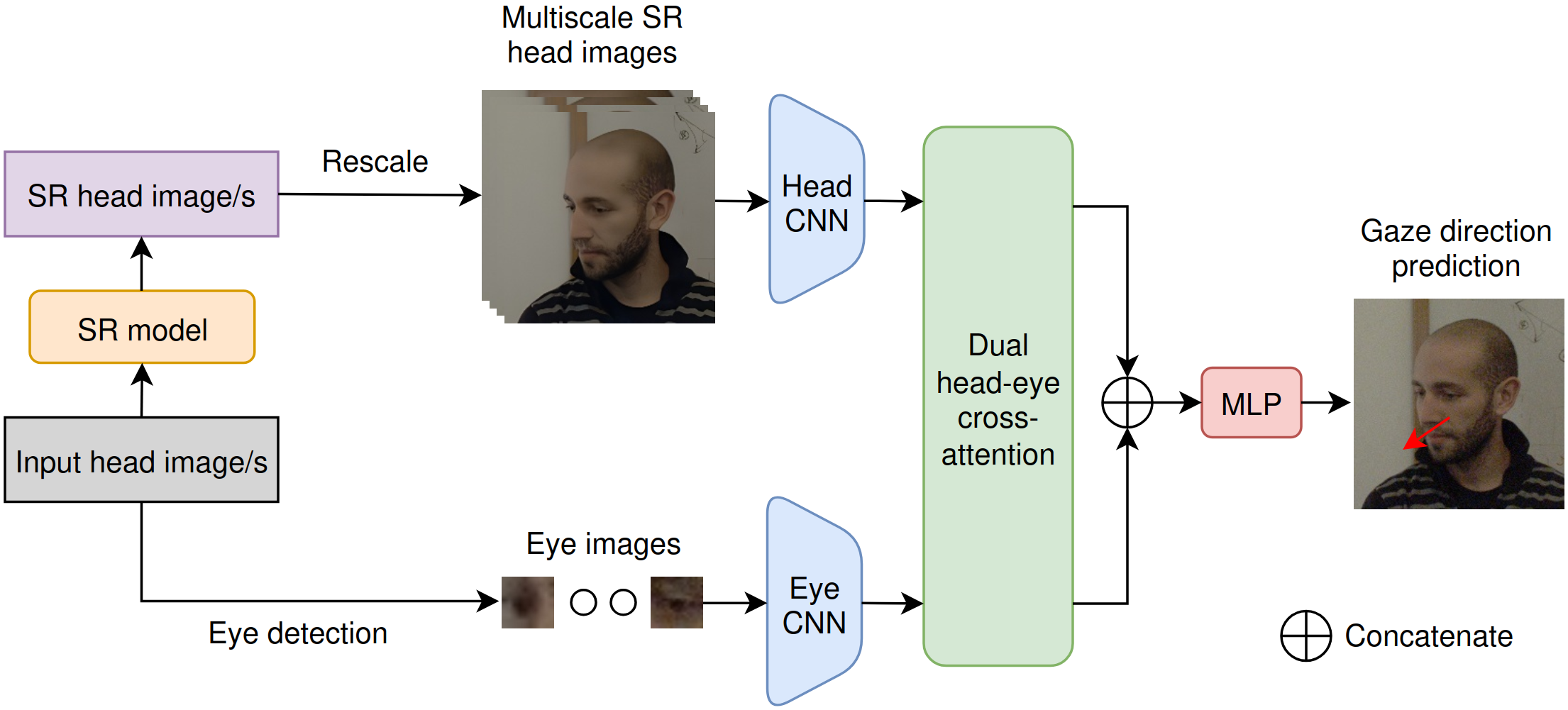}
\caption{The main idea behind DHECA-SuperGaze. The method leverages super-resolution (SR) and dual cross-feature attention between head and eye visual features to estimate the gaze direction.}
\label{fig1}
\end{figure}

In this paper, we introduce DHECA-SuperGaze, a deep learning-based method that improves the accuracy of gaze estimation by efficiently utilizing head and eye image features. The proposed method is illustrated in Fig. \ref{fig1}.
While traditional approaches usually model gaze estimation directly from head images, in \cite{head_eye_difference}, it is shown that there can be up to 35°  divergence between gaze and head alignment, making it necessary for methods to consider a close-up eye appearance in addition to head orientation.
Our architecture addresses this challenge through a hybrid convolutional-transformer design with a dual head-eye cross-attention (DHECA) module that processes eye and head image inputs in static and temporal (dynamic) settings. 
Furthermore, by integrating image super-resolution (SR) and multiscale processing into the method, DHECA-SuperGaze achieves state-of-the-art (SOTA) performance across Gaze360 \cite{gaze360} and GFIE \cite{gfie} datasets.

To the best of our knowledge, our contribution is threefold:
\begin{itemize}
    \item We address the wrongly annotated data in one of the most widely used gaze estimation datasets: Gaze360. We detected images with incorrect annotations and provide the corresponding corrected annotations. We find that Gaze360 rectification leads to better performance in all models that use these annotations.
    \item We present a novel DHECA module that leverages dual cross-feature attention between head and eye visual features. We demonstrate the importance of the proposed module in unconstrained gaze estimation.
    \item We introduce DHECA-SuperGaze, a novel method for static and temporal unconstrained gaze estimation from head and eye images. Our method achieves SOTA results on Gaze360 and GFIE datasets in both within- and cross-dataset evaluation.
\end{itemize}

\section{Preliminaries and related work}
\label{sec:preliminaries_related_work}

\subsection{Self/cross-attention}
The attention concept, popularized by a milestone paper \textit{Attention is all you need} \cite{vaswani}, refers to a mechanism that allows neural networks to focus on relevant information and enables the modeling of distant dependencies. Although originating from natural language processing, the concept was later popularized for computer vision applications in ViT \cite{dosovitskiy} paper. Within the attention mechanism, parts of the input data, known as tokens, are first projected to the query (Q), key (K), and value (V) matrices. Later, the Q and K-V pairs are mapped to an output by multiplication, scaling, and softmax activation. Self-attention is a type of attention in which Q, K, and V are projected from tokens that originate from the same source, e.g. from words of a particular sentence or patches of an image. On the other hand, cross-attention utilizes tokens from one source for Q, whereas tokens from the other source are taken for K-V pairs. These two sources can come from the same modality, such as images \cite{change_captioning_cross, crossgaze} and body joints \cite{body_joints_cross}, or even from different modalities, such as image and text \cite{image_cross_text, image_cross_text2}, RGB and thermal images \cite{rgb_thermal}, image and point cloud \cite{rgb_cross_pointcloud}, image and audio stream \cite{image_cross_audio}, etc.

\subsection{Super-resolution}

The goal of SR is to accurately restore the image resolution, i.e., to provide a faithful mapping from low to high resolution. With the rise of deep learning, Dong \textit{et al.} \cite{srcnn} proposed a convolutional neural network (CNN) for SR more than a decade ago, whereas Ledig \textit{et al.} \cite{srgan} later designed a generative adversarial network (GAN) for SR. Recently, attention-based SR models have gained increased popularity following the successful integration of transformers into image and video SR tasks \cite{ipt, vsr_transformer}. Particularly, Liang \textit{et al.} \cite{swinir} introduced an SR architecture based on the Swin transformer \cite{swin}, Hsu \textit{et al.} \cite{drct} extended the aforementioned SR architecture with dense connections, etc.

\subsection{Gaze estimation}

A decade ago, Zhang \textit{et al.} \cite{zhang15} proposed a model for gaze estimation from eye images that relied on the LeNet \cite{lenet} backbone. Later, Palmero \textit{et al.} \cite{palmero} designed a temporal gaze estimation model that utilized head and eye visual features produced by the VGG-16 \cite{vgg} backbone, while temporal modeling was achieved using a gated-rectified unit (GRU). In \cite{gaze360}, temporal modeling was performed through long short-term memory (LSTM) cells, while the model was optimized using a novel Pinball loss function, which incorporated the optimization of gaze prediction uncertainty into the training process.
The proposed model utilized the ResNet18 \cite{resnet} backbone, which was later followed by \cite{ashesh, gazetr, gazecaps, stage}. Furthermore, the proposed model was extended in \cite{ashesh} with the use of multiscale input, temporal modeling through max-pooling, and trigonometric prediction of gaze angles. 
Following increased interest in transformer-based gaze estimation, self-attention was used to model spatial dependencies in \cite{oh, gazetr, nagpure, mcgaze, stage} and temporal dependencies in \cite{mcgaze, stage}.

Recently, in \cite{crossgaze}, the authors proposed a gaze estimation method based on an eye-to-head cross-attention module. In particular, the head features extracted by the Inception-ResNet \cite{inception-resnet} backbone and the eye features obtained by the ResNet18 backbone were passed through a cross-attention module, where the eye features served for computing Q, whereas the head features served for computing K-V pairs. However, such a design does not exploit the full potential of cross-feature attention between head and eye features since it focuses solely on eye-to-head cross-attention.

\begin{figure*}
\centering
\begin{tabular}{ccc}
{\includegraphics[width = 0.64\columnwidth]{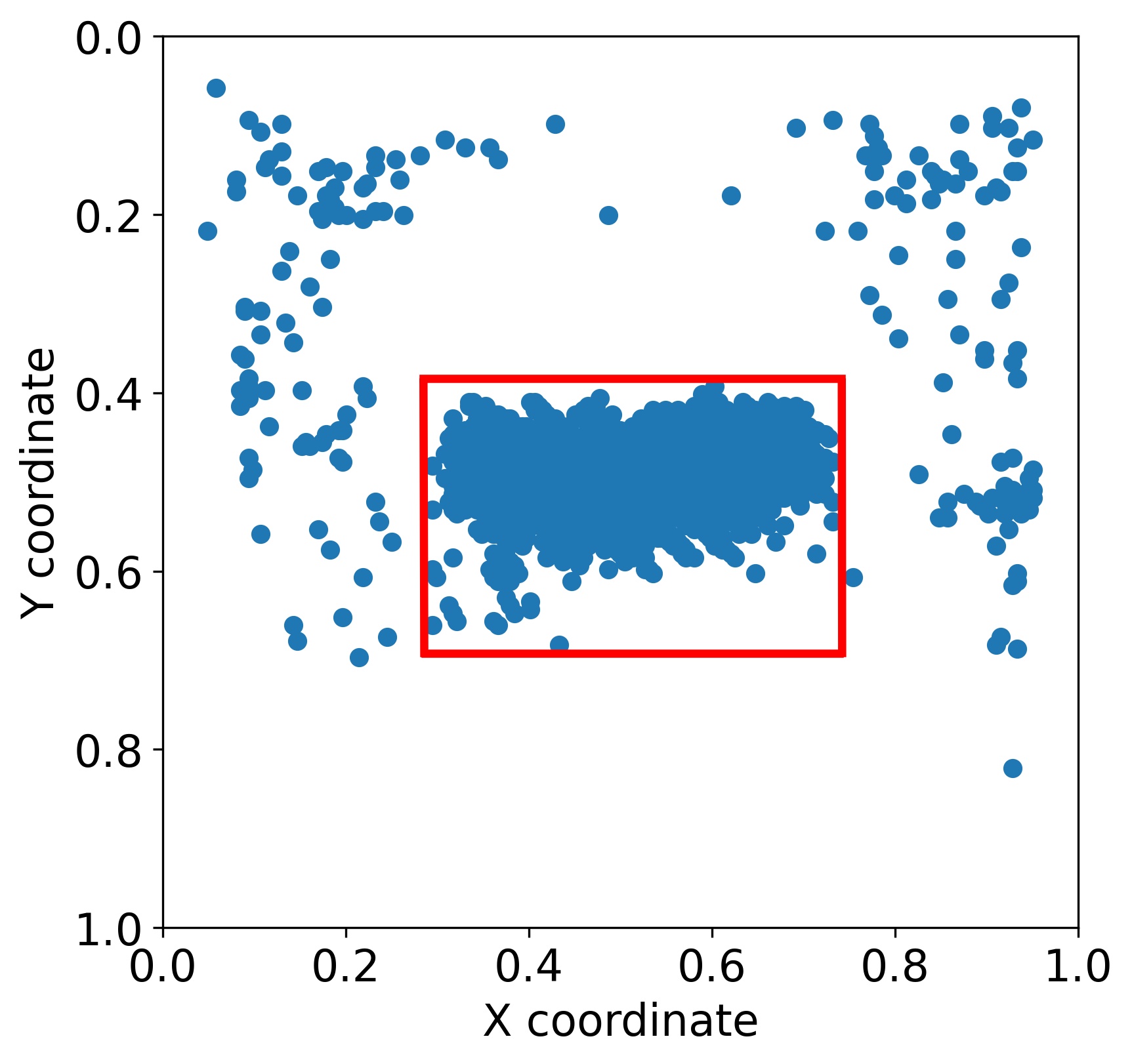}} &
{\includegraphics[width = 0.64\columnwidth]{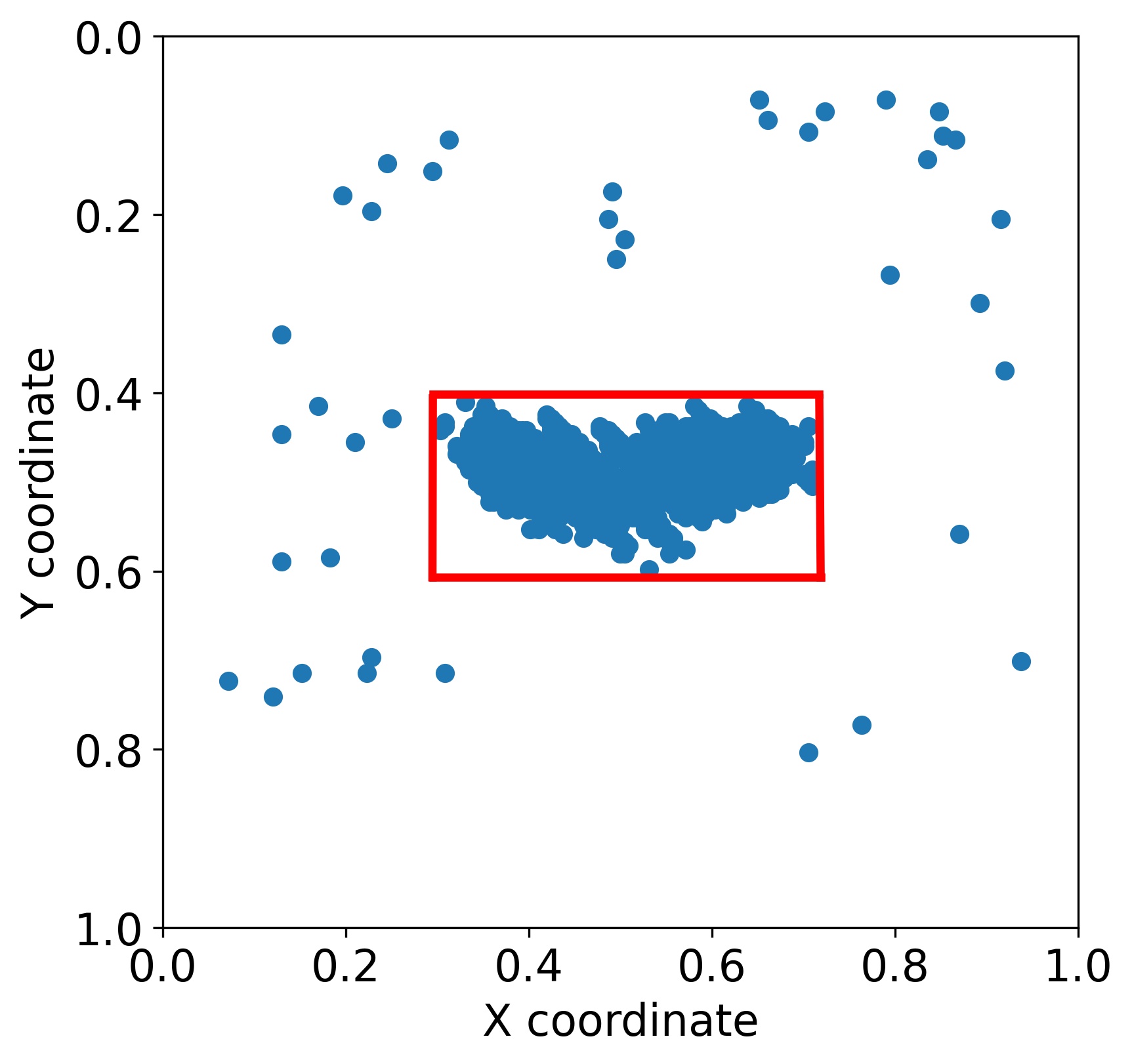}} &
{\includegraphics[width = 0.64\columnwidth]{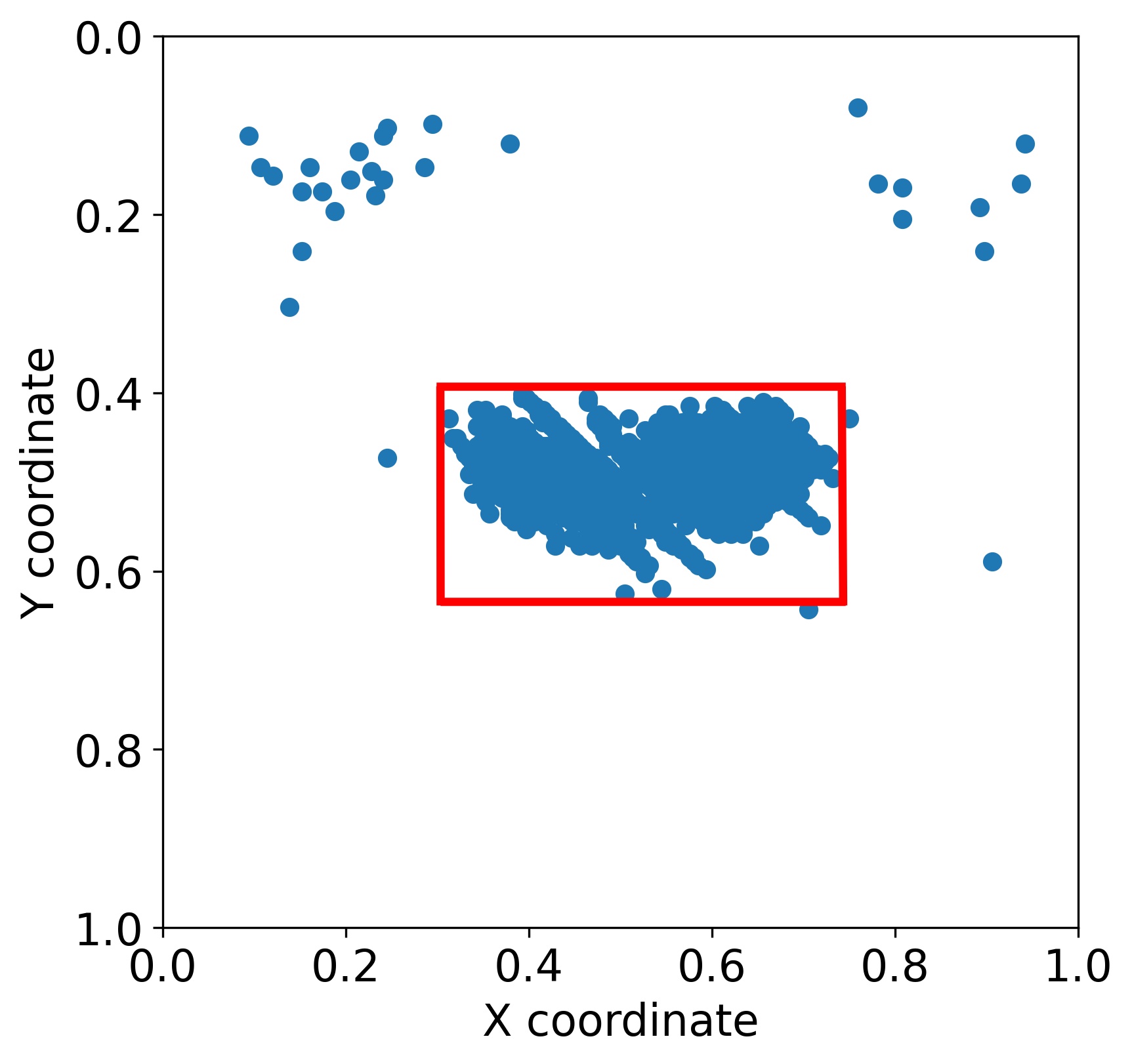}}
\end{tabular}
\caption{Distribution of unique face locations across images from the train (left), validation (center), and test (right) subsets of the Gaze360 dataset. Red rectangles encapsulate areas within which the annotations are valid.}
\label{fig_faces_position}
\end{figure*}

\begin{figure}
\centering
\includegraphics[width=\columnwidth]{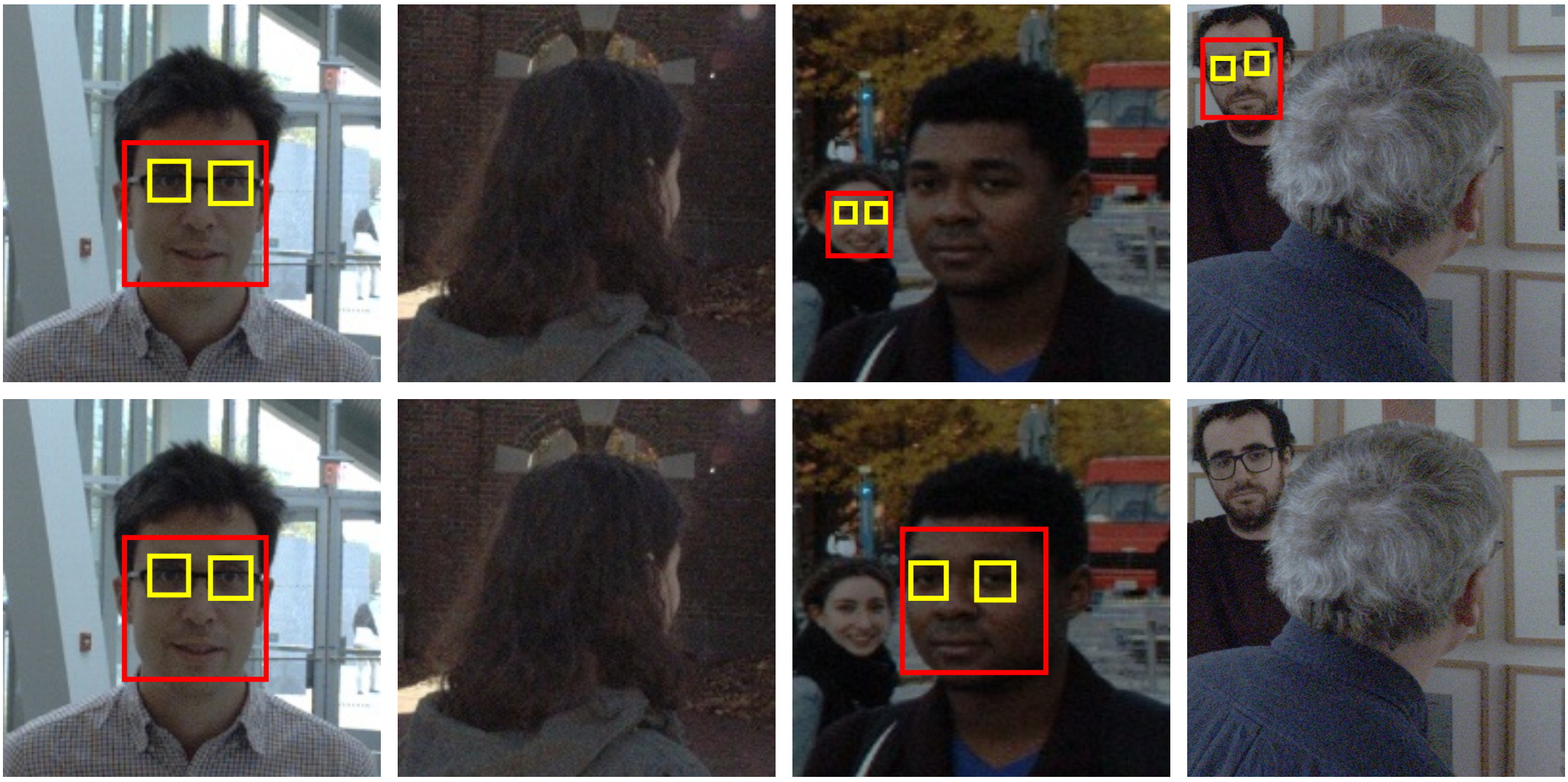}
\caption{Annotations from the original (top) and rectified (bottom) Gaze360 dataset. Red and yellow rectangles mark the position of the face and eyes, respectively. The original annotations were kept in the two left columns and rectified in the two right columns.}
\label{fig_gaze360_annotations}
\end{figure}

In \cite{haze-net, supervision}, gaze estimation methods that employ SR on head images were proposed. On the other hand, Wang \textit{et al.} \cite{gafuse-net} proposed a gaze estimation method that utilizes SR to restore the resolution of eye images. It is still unclear whether the optimal SR configuration includes applying SR to head images, eye images, or both head and eye images. In addition, the aforementioned gaze estimation methods that utilize SR were evaluated on MPIIGaze \cite{zhang15} and EyeDiap \cite{eyediap} datasets, some of the oldest datasets in the field, which include a very narrow distribution of frontal gaze vectors. Hence, an evaluation of the optimal SR configuration for gaze estimation on 360° datasets is yet to be performed.

\section{Datasets}
\label{sec:datasets}

In this research, we used two large gaze estimation datasets: GFIE and Gaze360. These datasets consist of in-the-wild video sequences. Additionally, both datasets cover a wide range of gazes across the possible 360° sphere.
The GFIE dataset was collected in various indoor locations. The acquired set of 71K frames was divided into 59K, 6K, and 6K training, validation, and test frames, respectively. The dataset displays a total of 61 subjects during a wide range of activities such as writing, moving objects, using electronic devices, etc.

Unlike GFIE, the Gaze360 dataset was acquired in both indoor and outdoor locations. In total, 172K frames were collected, split into 129K, 17K, and 26K training, validation, and test frames, respectively. The dataset contains 238 subjects that cover a wide diversity of age, ethnic, and gender groups. For each frame, a 3D gaze vector of a particular subject shown in the frame is given. In addition to gaze annotations, for frames in which a face was detected using \textit{dlib} \cite{dlib} library, bounding boxes that encapsulate the face, left eye, and right eye regions were provided.

\begin{figure*}
\centering
\includegraphics[width=\textwidth]{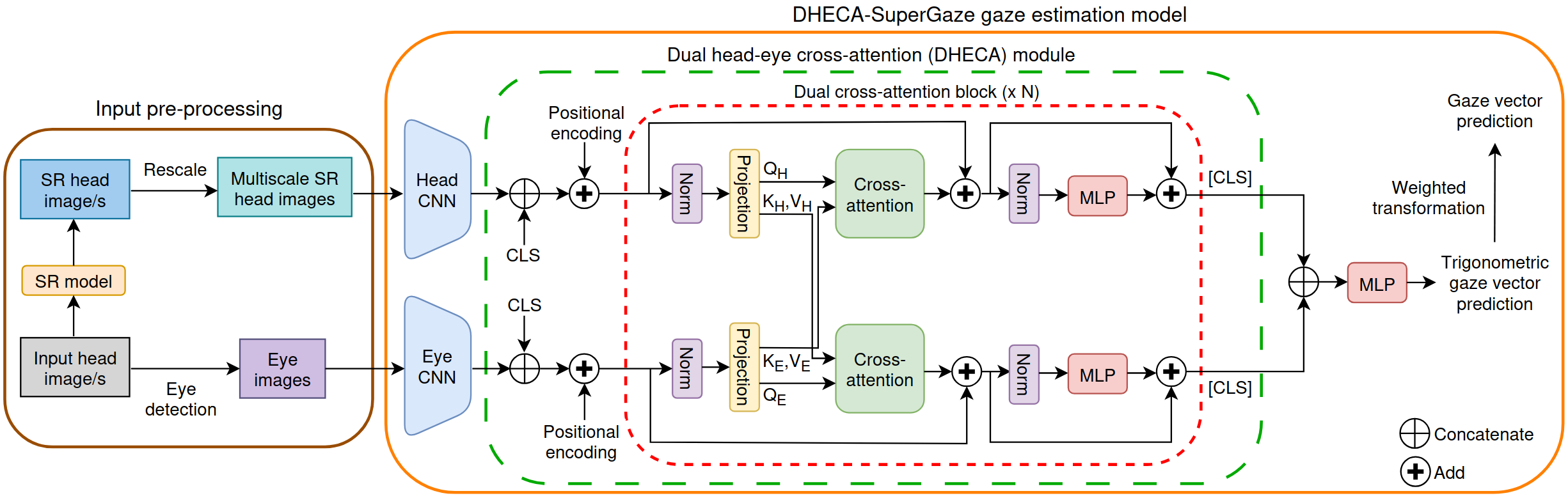}
\caption{Scheme of the proposed method, divided into input pre-processing (brown rounded rectangle) and gaze estimation model (orange rounded rectangle). Within the proposed model, the dual cross-attention block and the DHECA module are marked with red and green dashed rounded rectangles, respectively.}
\label{fig_method}
\end{figure*}

\begin{figure*}
\centering
\includegraphics[width=\textwidth]{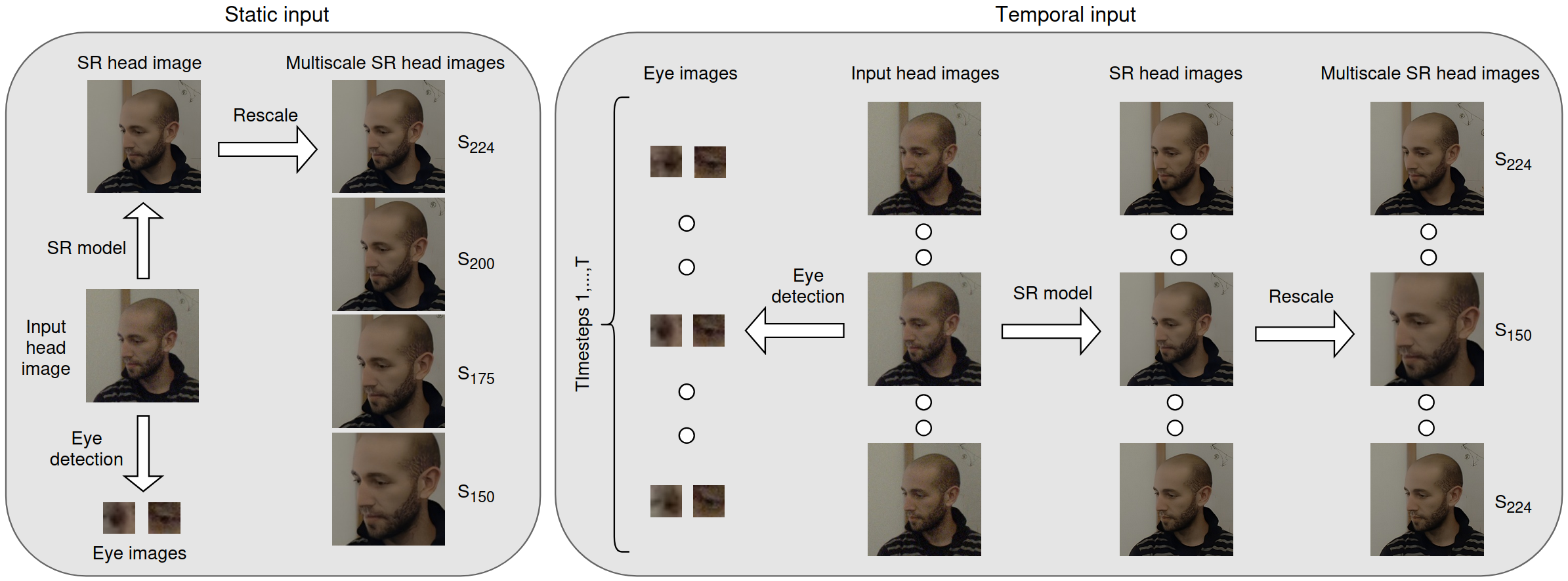}
\caption{Example of input pre-processing in static and temporal settings. In addition to applying eye detection to input head image/s, the input is passed through an SR model, whose result is eventually rescaled using particular scales.}
\label{fig_input}
\end{figure*}

During our exploratory data analysis of Gaze360, we detected a major issue with a part of the frame annotations. Particularly, for some frames we noticed that the provided bounding boxes encapsulate, instead of the face and eyes of the recorded subject (who is central in the frame and for whom the gaze vector is recorded), the face and eyes of other subjects present in the frame as illustrated in Fig. \ref{fig_gaze360_annotations}. Therefore, there is a partial misalignment between gaze annotations and face/eye bounding boxes.
To address this issue, we decided to find the incorrectly annotated images and rectify the corresponding annotations. For this purpose, we constructed face location distribution graphs, shown in Fig. \ref{fig_faces_position}, as follows. For each image in which \textit{dlib} localized a face and corresponding eyes, we simply calculated the normalized (image resolution-wise) center location of the detected face, and marked that location on the graph with a blue dot. Although it was easy to determine the general area of the obtained graphs that contains valid annotations, we needed to perform a visual inspection of some edge cases (samples whose location of the face center is close to the main cluster) to determine the exact valid intervals in each graph.
Finally, the following intervals of valid face locations for each subset were obtained:
\begin{itemize}
    \item Train - X: [0.29-0.74], Y: [0.39-0.69],
    \item Validation - X: [0.30-0.72], Y: [0.41-0.61],
    \item Test - X: [0.31-0.74], Y: [0.40-0.63].
\end{itemize}
Once we identified valid intervals, we rectified the annotations in the following manner. For each image that has a face annotation that falls outside the valid interval of its subset, we discarded its face and eye annotations and reapplied the \textit{dlib} face detector. If \textit{dlib} detected a face whose center falls within the valid interval, in addition to the already discarded face, (as it can in the second from the right column in Fig. \ref{fig_gaze360_annotations}), the locations of the detected face and the corresponding eyes were saved.

\section{Method}
\label{sec:method}

An overall scheme of the proposed method is illustrated in Fig. \ref{fig_method}. The method was inspired by several previous gaze estimation papers, namely \cite{ashesh, supervision, crossgaze}. In particular, the method presented in \cite{ashesh} served as the starting point, to which the use of eye images, SR, and the DHECA module was added. Generally, the method is divided into input pre-processing and prediction using a gaze estimation model.

\subsection{Pre-processing}

Fig. \ref{fig_input} displays an example of input pre-processing in both static and temporal settings.
First, we will explain the workflow of the method under static conditions, i.e., when a single head image is given as the input. The \textit{dlib} face landmark detector is applied to the head image to localize the positions of the eyes within the image. The obtained landmark positions are then used to crop eye images from the input image. Additionally, the input image is passed to the SR model to produce an enhanced version of the head image (SR head image). To perform the SR task, we utilized the SOTA SR model DRCT \cite{drct}. Particularly, we employed a GAN-based model of the DRCT method, Real-DRCT-GAN, which is provided in the official GitHub repository of the DRCT method \cite{drct_github}. The resulting SR head image is then resized to a $224\times224$ resolution and rescaled to obtain four multiscale head images as described in \cite{ashesh}. In particular, for each scale $S_i$ in $[S_i, i\in[224, 200, 175, 150]]$, the SR head image is center-cropped with the $i\times i$ window and resized to the original resolution.
% Jel ovo ispod potrebno?
For instance, for the scale $S_{150}$, a central $150\times150$ patch of $224\times224$ SR head image is cropped and resized back to $224\times224$.

For the temporal setting, i.e., the input consisting of head images from $T$ consecutive timesteps, eye detection and SR resolution steps are performed on the head image from each timestep. In line with \cite{ashesh}, we set $T$ to 7 and rescale the SR head images as follows. Each of the obtained seven consecutive SR head images is rescaled with its corresponding scale from $[S_i, i\in[224, 200, 175, 150, 175, 200, 224]]$, resulting in a zoom-in effect from the start of the sequence towards the central frame and a zoom-out effect from the central frame onward.

\subsection{Gaze estimation model}

Following pre-processing, visual features are extracted from multiscale SR head images and eye images by their respective CNN backbone. We utilize ResNet18 for both the head and eye backbones. Head CNN produces $H_H \times W_H \times C$ ($height \times width \times channel$) features for each image, which are then reshaped into $(H_H \times W_H) \times C$ tokens (i.e., $(H_H \times W_H)$ tokens of size $C$). Since static input to head CNN includes four multiscale SR head images, extraction of visual features results in $4 \times (H_H \times W_H) \times C$ tokens reshaped to $(4 \times H_H \times W_H) \times C$.
Similarly, the head CNN produces $(T \times H_H \times W_H) \times C$ tokens in temporal settings.
Moreover, the eye CNN extracts $H_E \times W_E \times C$ features for each eye image, resulting in $(2 \times H_E \times W_E) \times C$ tokens for static input and $(2 \times T \times H_E \times W_E) \times C$ tokens for temporal input.

Next, the extracted head and eye tokens are passed through the DHECA module as follows. A classification (CLS) token of length $C$ is first concatenated to head and eye tokens in their respective branches. Following the concatenation, positional encoding is added to head and eye tokens independently.
The obtained head and eye tokens are then processed by a dual cross-attention block. Particularly, head and eye tokens are normalized (layer-wise) and linearly projected into the $Q_H$, $K_H$, and $V_H$ matrices and the $Q_E$, $K_E$, and $V_E$ matrices, respectively. The cross-attention between head and eye features is performed as defined by the following formula:
\begin{equation}
    Cross\textbf{-}attention = Softmax(\frac{Q'K^T}{\sqrt{d_k}}) V,
\end{equation}
where $Q'$ comes from one branch (e.g., from the head branch, $Q_H$), $K$ and $V$ come from the other branch (e.g., from the eye branch, $K_E$ and $V_E$), whereas $d_k$ denotes the dimension of vectors of $K$. In each branch, the cross-attention output is then summed with the input of the dual cross-attention block. The result of the addition is then normalized (again layer-wise) and passed through a multilayer perceptron (MLP), whose output is later summed with the result of the previous addition. The output obtained from the head and eye branches is then passed to the next dual cross-attention block. Finally, after the $N$-th dual cross-attention block, classification tokens from the head and eye branches serve as the output of the DHECA module. The aforementioned classification tokens are then concatenated and processed by an MLP that provides the gaze prediction. In line with previous studies \cite{gaze360, ashesh}, in temporal settings, the gaze is predicted for the central frame of the input sequence.
Furthermore, once again following \cite{ashesh}, rather than directly predicting the gaze vector, we predict the following sine and cosine transforms of the yaw ($\theta$) and pitch ($\phi$) angles of the gaze vector: sine of yaw ($sy$), cosine of yaw ($cy$), and sine of pitch ($sp$). $\phi$ is simply obtained using $sin^{-1}(sp)$. On the other hand, $\theta$ is obtained through a two-step process as follows. First, sine- and cosine-based predictions of yaw ($\theta_S$ and $\theta_C$) are calculated by the following formulas:
\begin{equation}
    \theta_S = \begin{cases}
        sin^{-1} (sy) , & \text{if} \ sign(cy) = 1 \\
        sign(sy) \cdot \pi - sin^{-1}(sy), & \text{otherwise},
\end{cases}
\end{equation}
\begin{equation}
    \theta_C = sign(sy) \cdot cos^{-1} (cy),
\end{equation}
where the $sign$ function returns 1 for non-negative values, and -1 otherwise. In the second step, $\theta$ is obtained as a weighted average of $\theta_S$ and $\theta_C$ that favors sine-based prediction around $0\text{°}$ and cosine-based prediction around $\pm90\text{°}$:
\begin{equation}
\theta = w \cdot \theta_S + (1-w) \cdot \theta_C ; \,
    w = |cos(\frac{\theta_S + \theta_C}{2})|.
\end{equation}

\section{Experiments and results}
\label{sec:experimental_results}

\subsection{Implementation details}

\begin{table*}
\centering
\captionof{table}{Within-dataset evaluation of static (top) and temporal (bottom) models on Gaze360 ($\mathcal{D}_{Gaze360}$) and GFIE ($\mathcal{D}_{GFIE}$) datasets. For both static and temporal groups, the best and second-best results are bolded and underlined, respectively. Lower is better.}
\begin{tabular}{P{4.2cm}P{1.0cm}P{1.1cm}P{1.1cm}P{1.2cm}P{0.1cm}P{1.0cm}P{1.1cm}P{1.1cm}P{1.2cm}}
 \hline
 \multirow{3}{*}{Model} & \multicolumn{4}{c!}{$\mathcal{D}_{Gaze360}$} & & \multicolumn{4}{c!}{$\mathcal{D}_{GFIE}$} \\ \cline{2-5} \cline{7-10}
 & \multirow{2}{*}{Full} & \multirow{2}{*}{Front} & Front & \multirow{2}{*}{Backward} & & \multirow{2}{*}{Full} & \multirow{2}{*}{Front} & Front & \multirow{2}{*}{Backward} \\
 & & & facing & & & & & facing & \\ \hline
 Pinball Static \cite{gaze360} & 15.95 & 13.09 & 12.97 & 26.24 & & 19.23 & \underline{13.62} & 11.81 & 29.85 \\
 MSA \cite{ashesh} & 13.90 & 12.23 & 12.25 & \underline{19.90} & & 18.03 & 15.04 & 12.93 & 23.69 \\
 GazeTR-Hybrid \cite{gazetr} & 15.29 & 12.88 & 13.06 & 23.94 & & 21.12 & 17.24 & 14.17 & 28.45 \\
 L2CS-Net \cite{l2cs} & 15.81 & 13.12 & 13.14 & 25.49 & & 17.73 & 14.72 & 11.79 & \underline{23.44} \\
 CrossGaze \cite{crossgaze} & \underline{13.81} & \underline{11.97} & \underline{11.80} & 20.54 & & \underline{17.48} & 13.76 & \underline{11.75} & 25.38 \\
 \hdashline
 DHECA-SuperGaze-S (ours) & \textbf{13.33} & \textbf{11.67} & \textbf{11.45} & \textbf{19.29} & & \textbf{14.53} & \textbf{13.20} & \textbf{11.69} & \textbf{18.41} \\
 \hline
 Pinball LSTM \cite{gaze360} & 13.68 & 11.44 & 11.32 & 21.75 & & 18.65 & 13.82 & 11.63 & 27.81 \\
 MSA+Seq \cite{ashesh} & \underline{12.48} & \underline{10.68} & \underline{10.15} & \underline{18.97} & & \underline{16.79} & \underline{13.77} & \underline{11.54} & \underline{22.51} \\
 MCGaze \cite{mcgaze} & 13.01* & 10.99* & 10.62* & 20.22* & & 18.14 & 13.93 & 11.69 & 26.03 \\
 Hybrid-SAM+LSTM \cite{stage} & 17.02 & 13.28 & 12.38 & 30.50 & & 24.46 & 19.96 & 17.29 & 33.02 \\
 Hybrid-SAM+Tx \cite{stage} & 17.32 & 13.44 & 12.45 & 31.31 & & 24.39 & 18.90 & 14.81 & 34.83 \\
 \hdashline
 DHECA-SuperGaze-T (ours) & \textbf{11.89} & \textbf{10.23} & \textbf{9.71} & \textbf{17.93} & & \textbf{13.79} & \textbf{12.25} & \textbf{11.34} & \textbf{17.39} \\ \hline
 \multicolumn{9}{c!}{\footnotesize * Differs from \cite{sikic} as it was their only reported model that utilizes eye/face annotations. We re-evaluated it with the rectified dataset.} &
\end{tabular}
\label{table_results_within_dataset}
\end{table*}

We trained our models for 100 epochs using a batch size of 64 on a single NVIDIA RTX 3090 GPU. During training, the $L_1$ loss was computed for the trigonometric gaze prediction and optimized using the Adam \cite{adam} optimizer. The learning rate was initially set at 0.0001 and kept constant throughout the training process.
The resolution of the images passed to head and eye CNNs was set to $224 \times 224$ and $64 \times 64$, respectively. In the scenario where eyes were not detected in the input head image, completely zero-filled (i.e., black) images were passed to the eye CNN.
Both ResNet18 CNN backbones were initialized using transfer learning from the ImageNet \cite{imagenet} dataset. The depth of the DHECA module, i.e., the number of dual cross-attention blocks, was set to 4.
For the experiments on the Gaze360 dataset, all models were trained and evaluated on the rectified version of the dataset.

\subsection{Quantitative evaluation}

We follow the evaluation of \cite{sikic}, where an extensive evaluation methodology for a fair comparison of gaze estimation methods was proposed. In line with this work, we performed five stochastic training runs for each model and reported the average angular error (AE) on the test subset, with AE defined as:
\begin{equation}
    AE = cos^{-1} \frac{g \cdot \hat{g}}{||g|| \, ||\hat{g}||},
\end{equation}
where $g$ and $\hat g$ mark the ground truth and predicted gaze direction vectors, respectively. In addition to evaluating the models on the entire test set, the models were also evaluated on the front facing ($\theta \in [-20\text{°}, 20\text{°}]$), frontal ($\theta \in [-90\text{°}, 90\text{°}]$), and backward ($\theta \in ([-180\text{°}, -90\text{°}) \cup (90\text{°}, 180\text{°}] )$) subsets of the test set. 

\textbf{Within-dataset evaluation.} Within-dataset evaluation, a type of evaluation in which the model is trained and tested on the same dataset, was performed on Gaze360  and GFIE datasets. Table \ref{table_results_within_dataset} displays the results obtained by our models and provides a comparison to existing models. DHECA-SuperGaze-S and DHECA-SuperGaze-T denote static and temporal gaze estimation models obtained using the proposed method, respectively. The results show that the proposed models achieved a lower AE than any existing model. In particular, DHECA-SuperGaze-S model obtained 0.48° and 2.95° lower AE than the second-best performing static model across the entire (full) Gaze360 and GFIE datasets, respectively. Similarly, DHECA-SuperGaze-T outperformed the second-best performing temporal model by 0.59° on the entire Gaze360 dataset and by 3.00° on the entire GFIE dataset.

\textbf{Cross-dataset evaluation.} Following the within-dataset experiments, we performed cross-dataset evaluation, a type of evaluation where the model is trained on one dataset and tested on another dataset. Table \ref{table_results_cross_dataset} shows a quantitative cross-dataset comparison between the proposed and existing models, where $A \rightarrow{} B$ denotes training on dataset $A$ and testing on dataset $B$. The experimental results show that our models exhibited better generalization on unseen datasets than any other model. For cross-dataset testing on GFIE (i.e., when GFIE is the testing dataset), our static and temporal models achieved 4.30° and 3.99° lower AE across the entire dataset than previously published models, respectively. Furthermore, it should be noted that our models obtained significantly better ($\sim$10° lower) results than the MSA and MSA+Seq models that served as a starting point for DHECA-SuperGaze, revealing the impact of the DHECA module and SR on cross-dataset generalization. Furthermore, cross-dataset testing on Gaze360 revealed that our models surpassed the best existing models by 1.53° and 1.73° across the entire dataset under static and temporal conditions, respectively.

\begin{table*}
\centering
\captionof{table}{Cross-dataset evaluation of static (top) and temporal (bottom) models. The best and second-best results in each model group are bolded and underlined, respectively.}
\begin{tabular}{P{4.2cm}P{1.0cm}P{1.1cm}P{1.1cm}P{1.2cm}P{0.1cm}P{1.0cm}P{1.1cm}P{1.1cm}P{1.2cm}}
 \hline
 \multirow{3}{*}{Model} & \multicolumn{4}{c!}{$\mathcal{D}_{Gaze360} \xrightarrow{} \mathcal{D}_{GFIE}$} & & \multicolumn{4}{c!}{$\mathcal{D}_{GFIE} \xrightarrow{} \mathcal{D}_{Gaze360}$} \\ \cline{2-5} \cline{7-10}
 & \multirow{2}{*}{Full} & \multirow{2}{*}{Front} & Front & \multirow{2}{*}{Backward} & & \multirow{2}{*}{Full} & \multirow{2}{*}{Front} & Front & \multirow{2}{*}{Backward} \\
 & & & facing & & & & & facing & \\ \hline
 Pinball Static \cite{gaze360} & 29.77 & 26.80 & 33.65 & 35.40 & & 51.74 & 45.03 & 46.59 & 75.90 \\
 MSA \cite{ashesh} & 33.26 & 34.23 & 39.04 & \underline{31.42} & & 44.61 & 43.40 & \underline{43.94} & \underline{48.86} \\
 GazeTR-Hybrid \cite{gazetr} & 34.98 & 30.83 & 35.58 & 42.84 & & 57.17 & 56.92 & 59.15 & 58.06 \\
 L2CS-Net \cite{l2cs} & 28.14 & 24.97 & 26.12 & 34.89 & & 47.27 & \underline{43.10} & 45.81 & 62.27 \\
 CrossGaze \cite{crossgaze} & \underline{27.82} & \underline{24.63} & \underline{26.09} & 34.21 & & \underline{44.47} & 43.55 & 44.08 & 48.93 \\
 \hdashline
 DHECA-SuperGaze-S (ours) & \textbf{23.52} & \textbf{24.31} & \textbf{25.99} & \textbf{22.08} & & \textbf{42.94} & \textbf{41.09} & \textbf{43.06} & \textbf{48.71} \\
 \hline
 Pinball LSTM \cite{gaze360} & \underline{26.66} & \underline{25.61} & \underline{32.48} & \underline{28.59} & & 51.06 & 46.39 & 45.15 & 67.84 \\
 MSA+Seq \cite{ashesh} & 34.34 & 35.45 & 43.23 & 32.24 & & \underline{43.06} & \underline{41.52} & \underline{42.31} & \underline{48.60} \\
 MCGaze \cite{mcgaze} & 31.29 & 32.02 & 34.17 & 30.95 & & 49.78 & 46.11 & 44.87 & 60.51 \\
 Hybrid-SAM+LSTM \cite{stage} & 36.88 & 32.97 & 36.77 & 44.31 & & 57.54 & 54.33 & 52.26 & 69.04 \\
 Hybrid-SAM+Tx \cite{stage} & 39.95 & 37.03 & 42.62 & 45.51 & & 56.46 & 52.33 & 50.88 & 71.38 \\
 \hdashline
 DHECA-SuperGaze-T (ours) & \textbf{22.67} & \textbf{23.20} & \textbf{25.87} & \textbf{21.71} & & \textbf{41.33} & \textbf{39.62} & \textbf{42.27} & \textbf{47.82} \\ \hline
\end{tabular}
\label{table_results_cross_dataset}
\end{table*}

\begin{table*}
\centering
\captionof{table}{Ablation on attention module used in the proposed static (top) and temporal (bottom) models.}
\begin{tabular}{P{3.0cm}P{1.0cm}P{1.0cm}P{1.8cm}P{1.2cm}P{0.1cm}P{1.0cm}P{1.0cm}P{1.8cm}P{1.2cm}}
 \hline
 \multirow{2}{*}{Attention module} & \multicolumn{4}{c!}{$\mathcal{D}_{Gaze360}$} & & \multicolumn{4}{c!}{$\mathcal{D}_{GFIE}$} \\ \cline{2-5} \cline{7-10}
 & Full & Front & Front facing & Backward & & Full & Front & Front facing & Backward\\ \hline
 No attention & 13.64 & 12.02 & 11.87 & 19.78 & & 16.03 & 14.79 & 13.42 & 19.82 \\
 Self-attention & 13.60 & 11.93 & 11.77 & 19.83 & & 15.81 & 14.54 & 13.03 & 19.74 \\
 CrossGaze CA & 13.54 & 11.84 & 11.71 & 19.76 & & 15.44 & 14.13 & 12.68 & 19.64 \\
 \hdashline
 DHECA (ours) & \textbf{13.33} & \textbf{11.67} & \textbf{11.45} & \textbf{19.29} & & \textbf{14.53} & \textbf{13.20} & \textbf{11.69} & \textbf{18.41} \\ \hline
 No attention & 12.16 & 10.43 & 10.12 & 18.31 & & 15.04 & 13.37 & 12.81 & 18.54 \\
 Self-attention & 12.07 & 10.35 & 9.98 & 18.22 & & 14.67 & 12.91 & 12.36 & 18.21 \\
 CrossGaze CA & 12.02 & 10.31 & 9.91 & 18.17 & & 14.33 & 12.64 & 11.94 & 17.98 \\
 \hdashline
 DHECA (ours) & \textbf{11.89} & \textbf{10.23} & \textbf{9.71} & \textbf{17.93} & & \textbf{13.79} & \textbf{12.25} & \textbf{11.34} & \textbf{17.39} \\ \hline
\end{tabular}
\label{table_attention_ablation}
\end{table*}

\begin{table*}[!ht]
\centering
\captionof{table}{Ablation on SR configuration used in the proposed method in static (top) and temporal (bottom) settings.}
\begin{tabular}{P{3.0cm}P{1.0cm}P{1.0cm}P{1.8cm}P{1.2cm}P{0.1cm}P{1.0cm}P{1.0cm}P{1.8cm}P{1.2cm}}
 \hline
 \multirow{2}{*}{SR configuration} & \multicolumn{4}{c!}{$\mathcal{D}_{Gaze360}$} & & \multicolumn{4}{c!}{$\mathcal{D}_{GFIE}$} \\ \cline{2-5} \cline{7-10}
 & Full & Front & Front facing & Backward & & Full & Front & Front facing & Backward\\ \hline
 No SR & 13.52 & 11.85 & 11.67 & 19.56 & & 15.31 & 13.81 & 12.48 & 19.50 \\
 SR head & \textbf{13.33} & \textbf{11.67} & \textbf{11.45} & \textbf{19.29} & & \textbf{14.53} & \textbf{13.20} & \textbf{11.69} & \textbf{18.41} \\
 SR head (eye crops) & 13.38 & 11.70 & 11.51 & 19.44 & & 14.84 & 13.49 & 12.04 & 18.83 \\
 SR head\&eyes & 13.47 & 11.78 & 11.62 & 19.40 & & 15.14 & 13.63 & 12.14 & 19.04 \\
 \hline
 No SR & 12.13 & 10.44 & 10.01 & 18.24 & & 14.68 & 13.01 & 11.93 & 18.91 \\
 SR head & \textbf{11.89} & \textbf{10.23} & \textbf{9.71} & \textbf{17.93} & & \textbf{13.79} & \textbf{12.25} & \textbf{11.34} & \textbf{17.39} \\
 SR head (eye crops) & 11.92 & 10.26 & 9.76 & 18.05 & & 13.98 & 12.47 & 11.59 & 17.85 \\
 SR head\&eyes & 12.06 & 10.36 & 9.85 & 18.10 & & 14.38 & 12.86 & 11.65 & 18.21 \\ \hline
\end{tabular}
\label{table_SR_configuration_ablation}
\end{table*}

\textbf{Ablation studies.}
To demonstrate the effectiveness of the proposed DHECA module, we compared our module to the following alternatives displayed in Table \ref{table_attention_ablation}: \textit{No attention} - no attention is used, head and eye features are separately max- and GAP-pooled as in \cite{ashesh}, concatenated, and passed through MLP; \textit{Self-attention} - head and eye tokens are concatenated and passed through a basic ViT block with self-attention; \textit{CrossGaze CA} - cross-attention proposed in CrossGaze \cite{crossgaze} is used, with eye tokens serving for computing Q and head tokens for computing K-V pairs. The results show that the proposed attention module obtained the lowest AE on both datasets in both static and temporal settings. In general, cross-attention modules (CrossGaze CA and DHECA) lead to better performance compared to self-attention or no attention at all.

Furthermore, since our method utilizes head images and eye crops, there exist several ways how SR can be used within the solution. Therefore, we have tested the following configurations: \textit{No SR} - no SR is used within the method; \textit{SR head} - SR is employed only for head images, \textit{SR head (eye crops)} - SR is used for head images, while eye images are later cropped from these SR head images; \textit{SR head\&eyes} - SR is used separately on head and eye images. The results obtained with the aforementioned configurations are shown in Table \ref{table_SR_configuration_ablation}. Quantitative evaluation suggests that applying SR to head images and using original eye images leads to optimal performance. In addition, it should be noted that using any configuration that uses the SR model helps to increase the performance on both datasets for both static and temporal models, revealing the benefits of SR in gaze estimation.

Finally, to fully explore the problem of incorrectly annotated data in Gaze360, we trained models that utilize eye annotations using original data and using the improved data produced after Gaze360 rectification. The obtained results are provided in Table \ref{table_adjusted_dataset_ablation}. The evaluation showcases the importance of dataset rectification, as all models obtained approximately 0.15° lower AE when trained with the improved annotations.

\begin{table}
\centering
\captionof{table}{Ablation on using the original (top) and rectified (bottom) version of Gaze360 in methods that utilize eye annotations.}
\begin{tabular}{P{2.5cm}P{0.8cm}P{0.8cm}P{0.9cm}P{1.2cm}}
 \hline
 \multirow{2}{*}{Model} & \multirow{2}{*}{Full} & \multirow{2}{*}{Front} & Front & \multirow{2}{*}{Backward} \\
 & & & facing & \\ \hline
 CrossGaze & 13.97 & 12.14 & 11.95 & 20.68 \\
 MCGaze & 13.13 & 11.14 & 10.79 & 20.29 \\
 DHECA- & \multirow{2}{*}{13.48} & \multirow{2}{*}{11.88} & \multirow{2}{*}{11.71} & \multirow{2}{*}{19.36} \\
 SuperGaze-S & & & & \\
 DHECA- & \multirow{2}{*}{12.01} & \multirow{2}{*}{10.38} & \multirow{2}{*}{9.90} & \multirow{2}{*}{18.03} \\
 SuperGaze-T & & & & \\
 \hline
 CrossGaze & 13.81 & 11.97 & 11.80 & 20.54 \\
 MCGaze & 13.01 & 10.99 & 10.62 & 20.22 \\
 DHECA- & \multirow{2}{*}{13.33} & \multirow{2}{*}{11.67} & \multirow{2}{*}{11.45} & \multirow{2}{*}{19.29} \\
 SuperGaze-S & & & & \\
 DHECA- & \multirow{2}{*}{11.89} & \multirow{2}{*}{10.23} & \multirow{2}{*}{9.71} & \multirow{2}{*}{17.93} \\
 SuperGaze-T & & & & \\ \hline
\end{tabular}
\label{table_adjusted_dataset_ablation}
\end{table}

\subsection{Qualitative evaluation}

\begin{figure}
\centering
\includegraphics[width=\columnwidth]{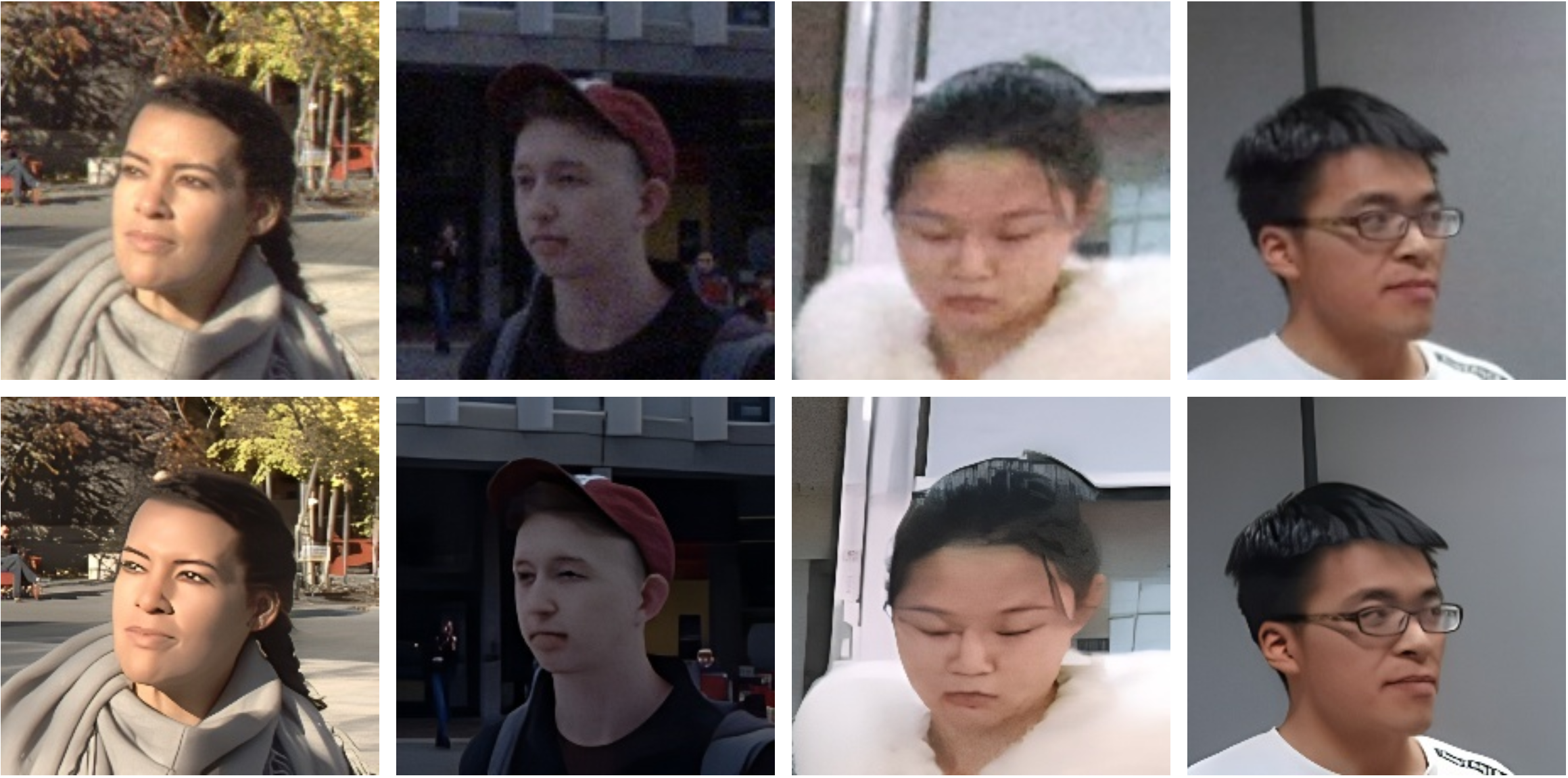}
\caption{SR results on images from Gaze360 (two left columns) and GFIE (two right columns) datasets. Original head images (top) were passed through the SR model to obtain SR head images (bottom). Best viewed zoomed in.}
\label{fig_qualitative_SR}
\end{figure}

\begin{figure}
\centering
\includegraphics[width=\columnwidth]{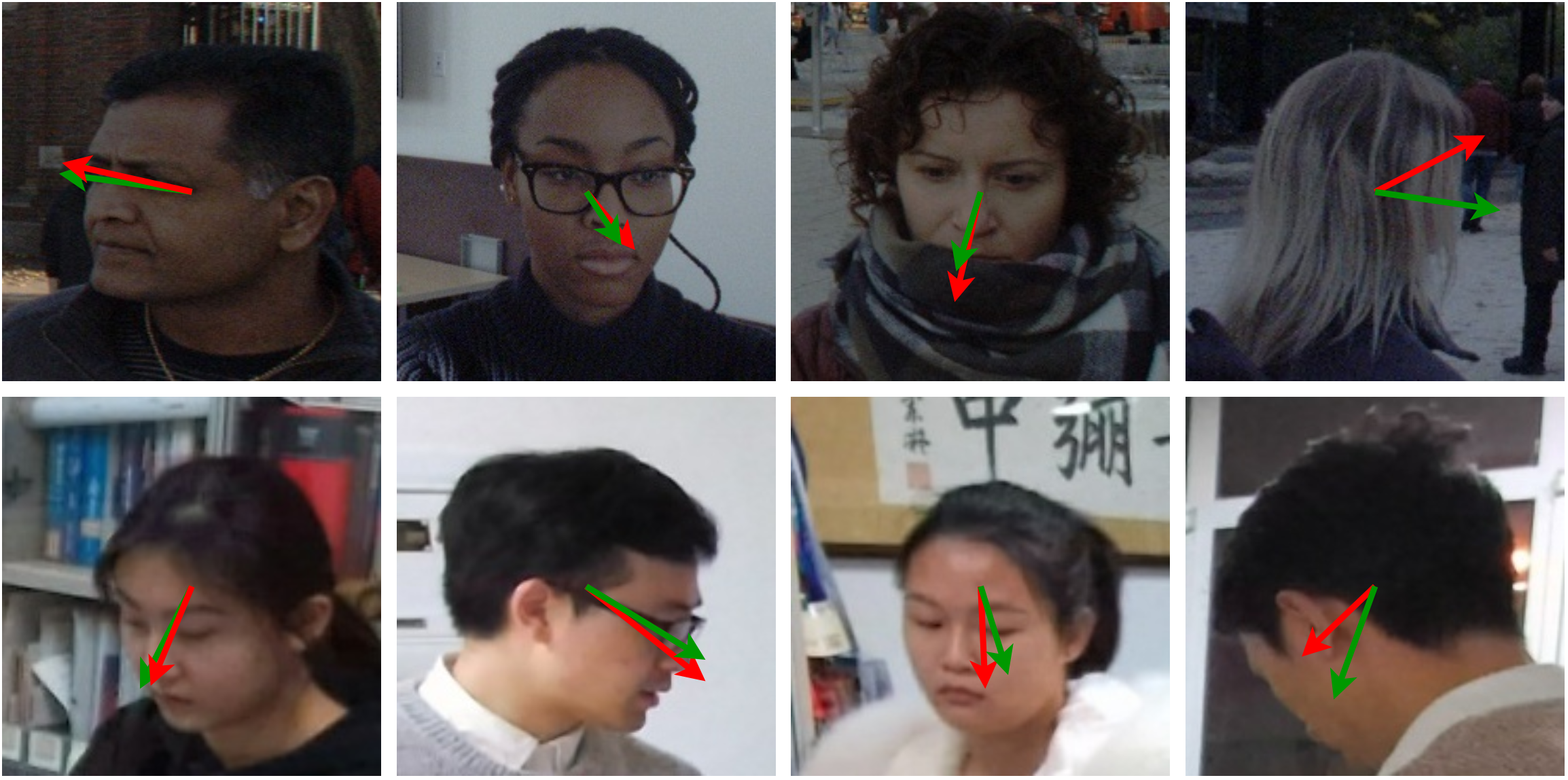}
\caption{Gaze prediction results on images from Gaze360 (top) and GFIE (bottom) datasets. Ground truth and predicted gaze vectors are depicted with green and red arrows, respectively.}
\label{fig_qualitative_gaze_prediction}
\end{figure}

In addition to quantitative evaluation, we performed a qualitative analysis to inspect the quality of outputs from the SR model and our gaze estimation model. Fig. \ref{fig_qualitative_SR} displays examples of original images from the Gaze360 and GFIE datasets, alongside the corresponding outputs from the SR model. The model managed to successfully increase the resolution of head images in various challenging scenes and illumination settings.
Moreover, Fig. \ref{fig_qualitative_gaze_prediction} illustrates gaze predictions of our static model in a wide range of Gaze360 and GFIE gazes. The model produced very accurate gaze direction vectors for frontal gazes, even in scenarios where a subject was wearing glasses. On the other hand, the estimation of backward gazes is much less accurate due to the unavailability of eye images. However, predicting such gazes is generally a more difficult task, even for humans.

\section{Conclusion}
\label{sec:conclusion}

In this paper, we have presented DHECA-SuperGaze, a method for gaze estimation in unconstrained environments that relies on SR and dual cross-feature attention between head and eye visual features.
The proposed method achieved excellent gaze estimation performance, surpassing existing methods in both within- and cross-dataset evaluation on Gaze360 and GFIE datasets.
The experimental results showed that the proposed DHECA module outperformed the alternatives under both static and temporal conditions, revealing the benefits of dual cross-feature attention.
In addition, the results show that the optimal SR configuration includes using original eye images and employing SR solely on head images.
Moreover, we have also detected and successfully overcome the problem of partially incorrect bounding boxes in Gaze360. Training on the rectified version of the dataset enabled all methods that utilize these annotations to achieve better results.
{
    \small
    \bibliographystyle{ieeenat_fullname}
    \bibliography{main}
}

\end{document}